\documentclass[conference]{IEEEtran}
\usepackage{times}
\usepackage{graphicx}
\usepackage{amsmath}
\usepackage{booktabs}
\usepackage[T1]{fontenc}
\usepackage[utf8]{inputenc}
\usepackage{natbib}
\usepackage{multirow}
\usepackage{caption}
\usepackage{url}
\usepackage{breakurl}
\usepackage{hyperref}

\begin{document}

\title{Aspect Extraction from E-Commerce Product and Service Reviews}

\author{
  {\normalfont
  \begin{tabular}{ccc}
    Valiant Lance D. Dionela & Fatima Kriselle S. Dy & Robin James M. Hombrebueno \\
    De La Salle University & De La Salle University & De La Salle University \\
    Manila, Philippines & Manila, Philippines & Manila, Philippines \\
    \small\texttt{valiant\_dionela@dlsu.edu.ph} & \small\texttt{fatima\_dy@dlsu.edu.ph} & \small\texttt{robin\_james\_hombrebueno@dlsu.edu.ph} \\[1em]
    
    Aaron Rae M. Nicolas & Charibeth K. Cheng & Raphael W. Gonda \\
    De La Salle University & De La Salle University & De La Salle University \\
    Manila, Philippines & Manila, Philippines & Manila, Philippines \\
    \small\texttt{aaron\_nicolas@dlsu.edu.ph} & \small\texttt{charibeth.cheng@dlsu.edu.ph} & \small\texttt{raphael.gonda@dlsu.edu.ph}
  \end{tabular}}
}

\maketitle

\begin{abstract}
Aspect Extraction (AE) is a key task in Aspect-Based Sentiment Analysis (ABSA), yet it remains difficult to apply in low-resource and code-switched contexts like Taglish, a mix of Tagalog and English commonly used in Filipino e-commerce reviews. This paper introduces a comprehensive AE pipeline designed for Taglish, combining rule-based, large language model (LLM)-based, and fine-tuning techniques to address both aspect identification and extraction. A Hierarchical Aspect Framework (HAF) is developed through multi-method topic modeling, along with a dual-mode tagging scheme for explicit and implicit aspects. For aspect identification, four distinct models are evaluated: a Rule-Based system, a Generative LLM (Gemini 2.0 Flash), and two Fine-Tuned Gemma-3 1B models trained on different datasets (Rule-Based vs. LLM-Annotated). Results indicate that the Generative LLM achieved the highest performance across all tasks (Macro F1 0.91), demonstrating superior capability in handling implicit aspects. In contrast, the fine-tuned models exhibited limited performance due to dataset imbalance and architectural capacity constraints. This work contributes a scalable and linguistically adaptive framework for enhancing ABSA in diverse, code-switched environments.
\end{abstract}

\section{Introduction}

In today's digital age, online platforms have emerged as the primary venues for public discourse, enabling individuals to share their opinions, thoughts, and experiences on a large scale \citep{Busst2024, Guzman2025}. This surge in user-generated content has led to the development of sentiment analysis, a field within Natural Language Processing (NLP) focused on the computational examination of opinions and emotions conveyed in text \citep{Nazir2022, Sagarino2022}. The applications of sentiment analysis are extensive, allowing businesses to automatically analyze customer reviews for market intelligence and enabling public health officials to assess public reactions to health crises \citep{Busst2024, Guzman2025}. According to \citet{Nazir2022}, sentiment analysis can be conducted on three primary levels of granularity: the document level, which assigns an overall sentiment to an entire text; the sentence level, which evaluates each sentence individually; and the aspect level, which provides the most nuanced insights by concentrating on specific features of a product or service.

This research centers on Aspect-Based Sentiment Analysis (ABSA), a fine-grained approach that addresses the shortcomings of general sentiment analysis by identifying opinions related to specific attributes \citep{Zhang2022}. A fundamental task within ABSA is Aspect Extraction (AE), which involves the automatic identification of particular characteristics of products or services being evaluated by consumers \citep{Santos2021, poria-etal-2014-rule}. For example, in a review that states, ``The laptop has an excellent screen but the battery is short-lived,'' an AE system would recognize ``screen'' and ``battery'' as the aspects under discussion \citep{Santos2021}. \citet{Zhang2022} emphasize that identifying these targets is the essential first step, as it defines what is being reviewed before any sentiment can be assigned. Successfully extracting these aspects equips businesses with structured data to make targeted improvements and helps consumers make more informed decisions.

Methodologies for aspect extraction have progressed to identify two main types of opinion targets: explicit aspects, which are directly stated terms like ``touchscreen'' or ``battery life,'' and the more complex implicit aspects, which must be inferred from context, such as ``affordable'' implying the ``price'' aspect \citep{poria-etal-2014-rule, Nazir2022, AL-Janabi2022}. Extracting implicit aspects presents considerable challenges, as it necessitates a profound level of natural language understanding to recognize underlying meanings \citep{AL-Janabi2022, Nazir2022}. Early methods for this task primarily employed rule-based approaches, relying on linguistic patterns and frequency analysis, but they often faced limitations in scalability \citep{poria-etal-2014-rule, Ali2020}. The field made significant strides with the introduction of deep learning models like LSTMs, which more effectively captured sequential dependencies in text \citep{Busst2024, Zhang2022}. More recently, the advent of transformer-based architectures, such as BERT, has transformed the landscape by achieving state-of-the-art results through the generation of deep, contextualized word embeddings that resolve ambiguity across various domains \citep{Santos2021, Busst2024}. However, a major drawback of these advanced supervised methods is their heavy dependence on large volumes of labeled training data, which can often be expensive and time-consuming to produce \citep{Santos2021}.

Despite the success of advanced models, a significant research gap remains in their application to low-resource languages and complex, code-switched linguistic environments. This challenge is particularly apparent in the Philippines, where online reviews are often composed in ``Taglish,'' a fluid blend of Tagalog and English \citep{Sagarino2022}. Current approaches that translate such text into a single language prior to analysis risk losing vital linguistic nuances and cultural context, which can lead to inaccurate outcomes \citep{Ali2020, Sagarino2022}. Furthermore, \citet{Andoy2022} highlight that Filipino consumers frequently express skepticism and have trust issues regarding the credibility of online reviews. This underscores a pressing issue: the lack of robust aspect extraction methods specifically tailored to navigate the complexities of code-switched Taglish, which is essential for accurately representing the authentic voice of Filipino consumers in their product and service reviews.

\section{Methodology}

\subsection{Dataset}
The study utilizes the \textit{SentiTaglishProductsAndServices} dataset \cite{ccosme2023sentitaglish}, consisting of 10,510 reviews collected from Google Maps and Shopee Philippines. Originally annotated only for sentiment polarity, a secondary annotation pass was performed to label General and Specific aspects.

\subsection{Topic Modeling and Framework Development}
To ensure aspect labels were data-driven rather than arbitrary, a multi-method approach (Methodological Triangulation) was employed to synthesize categories:

\begin{enumerate}
    \item \textbf{Open Coding:} A qualitative analysis of data subsets was conducted to identify themes directly from the text without pre-existing constraints.
    \item \textbf{LDA (Latent Dirichlet Allocation):} Multiple LDA models were trained, with the optimal model (Coherence Score: 0.5389) selected. While useful, LDA topics often required significant manual interpretation (e.g., mixing "gumagana" and "complete").
    \item \textbf{BERTopic:} Multilingual sentence embeddings were utilized to generate semantically coherent topics. This method successfully isolated granular themes such as specific product defects and sensory feedback (e.g., "amoy", "mabango").
    \item \textbf{Generative LLM Consultation:} Large Language Models (Gemini and Grok) served as research assistants to synthesize the outputs of LDA and BERTopic into a logical hierarchy.
\end{enumerate}

The resulting \textbf{Hierarchical Aspect Framework (HAF)} organizes aspects into four General Categories---\emph{Product, Delivery, Price,} and \emph{Service}---branching into 21 granular Specific Aspects (Table \ref{tab:haf}).

\begin{table}[h]
\centering
\small
\resizebox{\columnwidth}{!}{%
\begin{tabular}{l|l}
\toprule
\textbf{General Aspect} & \textbf{Specific Aspects} \\
\midrule
\textbf{PRODUCT} & Color, Condition, Correctness, Durability, \\
& Effectiveness, Functionality, Material, \\
& Sensory, Size/Measurement, General \\
\hline
\textbf{DELIVERY} & Condition, Correctness, Timeliness, General \\
\hline
\textbf{PRICE} & Affordability, Value for Money, General \\
\hline
\textbf{SERVICE} & Handling, Responsiveness, Trustworthiness, \\
& General \\
\bottomrule
\end{tabular}
}
\caption{The Hierarchical Aspect Framework (HAF).}
\label{tab:haf}
\end{table}

\subsection{Data Annotation Pipeline}
A semi-automated annotation pipeline was implemented. First, a calibration subset of 67 reviews was manually annotated by four researchers. After five iterative rounds of guideline refinement, a Fleiss' Kappa of \textbf{0.691} (Good Agreement) was achieved.

Using this gold standard as a few-shot prompt, Gemini 2.0 Flash was employed to annotate the remaining dataset. To validate this semi-automated process, random samples were manually inspected. For General Aspects, the LLM achieved an accuracy of 92.31\% by the seventh round of prompt refinement, confirming the reliability of the generated dataset (LLM-DS) for training downstream models.

\subsection{Aspect Identification Models}
Four distinct approaches were implemented to compare different paradigms of NLP.

\subsubsection{Rule-Based System}
A custom rule-based tagger was developed operating on two levels:
\begin{enumerate}
    \item[(a)] \textbf{Foundational Rules:} Utilized Regex matching and a keyword lexicon based on the codebook, which was expanded using high-probability terms identified via BERTopic analysis to improve coverage (e.g., matching ``mura'' to \emph{Price-Affordability}).
    \item[(b)] \textbf{Contextual Disambiguation:} Ten specific linguistic rules were applied to handle ambiguity. For instance, the word "bilis" (fast) could refer to \emph{Delivery-Timeliness} or \emph{Product-Effectiveness}. The system checks surrounding context (e.g., "dumating" vs. "epekto") to assign the correct tag.
\end{enumerate}

\subsubsection{Large Language Model (LLM)}
Google's Gemini 2.0 Flash was utilized via API with few-shot prompting. The model processed raw review text using a structured prompt containing HAF definitions. It was tasked with multi-label classification, outputting boolean vectors indicating present aspects.

\subsubsection{Fine-Tuned Gemma Models}
Two separate fine-tuning experiments were conducted using the \texttt{gemma-3-1b-pt} model \citep{gemma_2025}. A Parameter-Efficient Fine-Tuning (PEFT) strategy with Low-Rank Adaptation (LoRA) was used (Rank=8, Alpha=16, Dropout=0.05). To address class imbalance (e.g., \emph{Service} aspects appearing in only 21\% of data), inverse frequency class weights were applied to the Binary Cross-Entropy loss function.

The two variants differed by their training data:
\begin{enumerate}
    \item \textbf{FT-Gemma (RB-DS):} Trained on the dataset annotated by the Rule-Based system. This dataset is "clean" but limited to explicit keywords.
    \item \textbf{FT-Gemma (LLM-DS):} Trained on the dataset annotated by Gemini. This dataset contains more implicit and nuanced labels.
\end{enumerate}

\begin{figure}[h]
  \centering
  \includegraphics[width=0.5\textwidth]{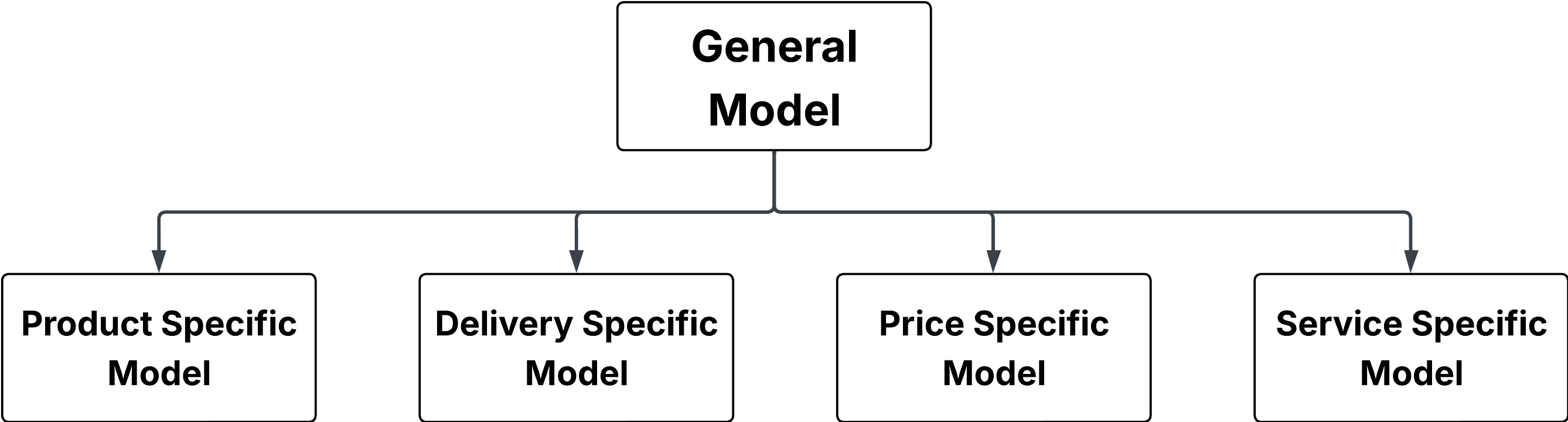} 
  \caption{Fine-Tuning Hierarchical Architecture}
  \label{fig:myimage}
\end{figure}

The training process followed a simple two-stage setup. The general model was trained first so it could learn how to identify the general aspects of a review. Once this model was established, the specific models were trained separately for each domain. Each specific model focused only on the labels that belonged to its own category. This setup allowed the system to avoid training one large model for all labels at once and instead rely on smaller models that were trained to handle only the details of their assigned aspect.

After the general and specific fine-tuning, the system transitions from a flat prediction setup to a hierarchical architecture. In the flat setup, each aspect classifier worked on its own and made predictions independently. In the hierarchical setup, the process begins with the general model, which produces the initial general aspect predictions. These outputs act as the conditions that determine which specific models will run in the next stage. If a general aspect is not predicted in the first stage, the system no longer proceeds with the specific aspects under that domain. This creates a clear decision flow where only the relevant specific models are activated, allowing the general model to handle broad patterns while the specific models focus on their own detailed labels.

\section{Results}

All models were evaluated on a held-out test set of manually annotated reviews. Performance was measured using Exact Match Accuracy (EM), Hamming Loss, and F1-Score.

\subsection{General Aspect Identification}
Table \ref{tab:general_results} presents the comparative performance. The LLM significantly outperformed all other methods (EM 78.26\%, Macro F1 0.91). The Rule-Based system served as a strong baseline (EM 52.17\%). Both Fine-Tuned models struggled significantly, with the model trained on the LLM dataset actually performing worse in Exact Match (8.7\%) than the one trained on rules (13.04\%).

\begin{table}[h]
\centering
\small
\resizebox{\columnwidth}{!}{%
\begin{tabular}{l|c|c|c|c}
\toprule
\textbf{Metric} & \textbf{LLM} & \textbf{Rule-Based} & \textbf{FT-LLM DS} & \textbf{FT-RB DS} \\
\midrule
Exact Match & \textbf{0.7826} & 0.5217 & 0.0870 & 0.1304 \\
Hamming Loss & \textbf{0.0761} & 0.1630 & 0.5543 & 0.4783 \\
Macro F1 & \textbf{0.9136} & 0.8200 & 0.6047 & 0.6452 \\
Micro F1 & \textbf{0.9243} & 0.8400 & 0.5888 & 0.6282 \\
\bottomrule
\end{tabular}
}
\caption{Comparative Performance on General Aspect Identification. FT-LLM DS refers to Gemma fine-tuned on the Gemini-annotated dataset; FT-RB DS refers to Gemma fine-tuned on the Rule-Based dataset.}
\label{tab:general_results}
\end{table}

\subsection{Specific Aspect Identification}
The performance gap widened when identifying granular specific aspects. Table \ref{tab:specific_results} provides a detailed breakdown of F1 scores across the four categories.

The LLM maintained robustness across all categories, particularly in \textbf{Delivery} (0.9474) and \textbf{Price} (0.8750). The Rule-Based model performed competitively in \textbf{Price} (0.5495) due to explicit keywords but dropped in performance for \textbf{Product} and \textbf{Service}. Both fine-tuned models exhibited catastrophic failure in the \textbf{Price} category, achieving an F1 score of 0.00, indicating an inability to learn even straightforward explicit patterns.

\subsection{Aspect Extraction Results}
Following identification, the LLM was tasked with extracting the exact text spans.

\begin{center}
\small
\begin{tabular}{lc}
\toprule
\textbf{Aspect Category} & \textbf{Token-Level F1} \\
\midrule
Price & 1.000 \\
Service & 0.869 \\
Delivery & 0.721 \\
Product & 0.627 \\
\bottomrule
\end{tabular}
\captionof{table}{Token-Level F1 Scores for Aspect Extraction using Gemini.}
\label{tab:extraction}
\end{center}

As shown in Table \ref{tab:extraction}, extraction was perfect for \emph{Price}, which relies on distinct numerical values. \emph{Product} extraction had the lowest F1 (0.627), often due to "Over-Extraction" errors where the model included unnecessary context words, or "Partial Matches" caused by Taglish variability.

\begin{table*}[t]
\centering
\small
\begin{tabular}{l|c|c|c|c|c|c|c}
\toprule
\textbf{Category} & \textbf{Metric} & \textbf{LLM} & \textbf{Rule-Based} & \textbf{FT-LLM DS (F)} & \textbf{FT-RB DS (F)} & \textbf{FT-LLM DS (H)} & \textbf{FT-RB DS (H)} \\
\midrule
\multirow{3}{*}{\textbf{Product}} & Precision & \textbf{0.6923} & 0.5882 & 0.1194 & 0.0688 & 0.1565 & 0.0957 \\
& Recall & \textbf{0.6923} & 0.3846 & 0.6000 & 0.4100 & \textbf{0.6923} & 0.4231 \\
& F1 & \textbf{0.6923} & 0.4651 & 0.1747 & 0.1173 & 0.2553 & 0.3123 \\
\hline
\multirow{3}{*}{\textbf{Delivery}} & Precision & \textbf{1.0000} & 0.5517 & 0.0250 & 0.0278 & 0.0455 & 0.0156 \\
& Recall & \textbf{0.9000} & 0.4103 & 0.2500 & 0.2000 & 0.1000 & 0.1000 \\
& F1 & \textbf{0.9474} & 0.4706 & 0.0455 & 0.0500 & 0.0626 & 0.0270 \\
\hline
\multirow{3}{*}{\textbf{Price}} & Precision & \textbf{0.8750} & 0.5952 & 0.0000 & 0.0000 & 0.0000 & 0.0000 \\
& Recall & \textbf{0.8750} & 0.5102 & 0.0000 & 0.0000 & 0.0000 & 0.0000 \\
& F1 & \textbf{0.8750} & 0.5495 & 0.0000 & 0.0000 & 0.0000 & 0.0000 \\
\hline
\multirow{3}{*}{\textbf{Service}} & Precision & \textbf{0.6667} & 0.5769 & 0.0625 & 0.1250 & 0.0457 & 0.0455 \\
& Recall & \textbf{0.8889} & 0.5172 & 0.2500 & 0.1250 & 0.2222 & 0.1111 \\
& F1 & \textbf{0.7619} & 0.5455 & 0.1000 & 0.1250 & 0.0728 & 0.0646 \\
\bottomrule
\end{tabular}
\caption{Detailed performance comparison for Specific Aspect Identification, including both Flat (F) and Hierarchical (H) Fine-Tuned Gemma Models.}
\label{tab:specific_results}
\end{table*}

\section{Discussion and Error Analysis}

The analysis reveals critical insights into the interplay between model architecture, data complexity, and linguistic nuance.

\subsection{Implicit vs. Explicit Aspects}
The performance gap is largely explained by the ability to handle implicit language. The Rule-Based system is limited to explicit keywords. For example, it successfully identifies \emph{Price} via keywords like "mura" (cheap). However, it fails to identify \emph{Delivery-Timeliness} in the phrase "It took a week to arrive" because the explicit keyword "late" is absent. The LLM excels here, correctly inferring the aspect from the context of time duration.

\subsection{Cross-Aspect Contamination}
A major challenge identified was the semantic overlap between \emph{Product} and \emph{Delivery} categories, which both share a "Correctness" sub-aspect.
\begin{quote}
    \textit{"Blue order ko pero pink dumating."} (I ordered Blue but Pink arrived.)
\end{quote}

This represents a fulfillment error (\emph{Delivery-Correctness}). However, the Fine-Tuned models frequently misclassified this as a product attribute flaw (\emph{Product-Correctness}). The LLM, leveraging broader semantic pre-training, successfully distinguished that the error lay in the transaction rather than the item's inherent quality.

\subsection{Architectural Failure of Fine-Tuning}
The failure of both Fine-Tuned Gemma 1B models—regardless of the training dataset—points to a "capacity mismatch."

\begin{enumerate}
    \item \textbf{FT-Gemma (RB-DS):} This model was trained on simple, explicit rule-based data. Consequently, it learned an overly simplistic "worldview" and failed to generalize to the nuanced, implicit language found in the actual test reviews.
    \item \textbf{FT-Gemma (LLM-DS):} This model was trained on the superior, implicit-aware dataset. However, this increased data complexity proved detrimental. The 1B parameter architecture likely lacked the capacity to encode the noisy, code-switched patterns of Taglish while simultaneously handling a multi-label classification task with 25 distinct labels. The model converged to a state of over-prediction (high recall, near-zero precision), effectively "spamming" labels rather than learning discriminative features.
\end{enumerate}

This highlights a critical finding: superior training data does not guarantee a superior model if the architecture lacks the parameters to effectively learn from it.

\section{Conclusion}

This study contributed to making a comprehensive baseline for Aspect Extraction in Taglish reviews. A Hierarchical Aspect Framework (HAF) was developed through methodological triangulation, providing a structured taxonomy for future research.

The comparative evaluation demonstrates that \textbf{Generative LLMs (Gemini 2.0 Flash)} provide the most robust solution for this low-resource, code-switched domain. It achieved the highest accuracy and F1 scores, effectively balancing explicit and implicit aspect identification. While \textbf{Rule-Based} systems offer computational efficiency for simple tasks, they lack the recall required for nuanced reviews. Conversely, fine-tuning smaller language models (1B parameters) proved ineffective given the high complexity and class imbalance of the dataset.

Future work should aim to expand the dataset to address imbalance, particularly for underrepresented Service aspects, and explore hybrid pipelines that combine the high precision of rules with the semantic reasoning of LLMs to optimize the trade-off between accuracy and computational cost.

\bibliographystyle{acl_natbib}
\bibliography{custom}

\end{document}